\documentclass[conference]{IEEEtran}
\IEEEoverridecommandlockouts
\usepackage{cite}
\usepackage{amsmath,amssymb,amsfonts}
\usepackage{algorithmic}
\usepackage{graphicx}
\usepackage{textcomp}
\usepackage{xcolor}
\usepackage{multirow} 
\usepackage{microtype}
\usepackage{subcaption}
\usepackage{tabularx}
\usepackage{lipsum}
\usepackage{latexsym}
\usepackage{times}

\def\BibTeX{{\rm B\kern-.05em{\sc i\kern-.025em b}\kern-.08em
    T\kern-.1667em\lower.7ex\hbox{E}\kern-.125emX}}
\begin{document}

\title{Few-shot Name Entity Recognition on StackOverflow\\
}

\author{
\IEEEauthorblockN{Xinwei Chen}
\IEEEauthorblockA{\textit{Department of Electronical and Computer Engineering} \\
\textit{University of Illinois at Urbana Champaign}\\
Urbana, USA\\
xinweic2@illinois.edu }
\and

\IEEEauthorblockN{Kun Li}
\IEEEauthorblockA{\textit{Department of Computer Science} \\
\textit{University of Illinois at Urbana Champaign}\\
Urbana, USA \\
kunli3@illinois.edu}

\and

\IEEEauthorblockN{Jiangjian Guo}
\IEEEauthorblockA{\textit{Department of Computer Science and Engineering} \\
\textit{University of California San Diego}\\
La Jolla, US \\
j9guo@ucsd.edu }

\and

\IEEEauthorblockN{Tianyou Song}
\IEEEauthorblockA{\textit{Department of Computer Science} \\
\textit{Columbia University}\\
New York, USA \\
tianyou.song@columbia.edu}
}

\maketitle

\begin{abstract}
StackOverflow, with its vast question repository and limited labeled examples, raise an annotation challenge for us. We address this gap by proposing RoBERTa+MAML, a few-shot named entity recognition (NER) method leveraging meta-learning. Our approach, evaluated on the StackOverflow NER corpus (27 entity types), achieves a 5\% F1 score improvement over the baseline. We improved the results further domain-specific phrase processing enhance results.
\end{abstract}


\section{Introduction}
The increasement of programming content on the internet presents challenges in understanding and extracting software related information. StackOverflow, as the largest programming forum, has over 15 million software related questions. To comprehend this vast corpus effectively, identifying named entities (NEs) is crucial. However, fully supervised learning for named entities recognition (NER) in domain-specific contexts demands extensive labeled data, which is resource-intensive. In response, we propose a few-shot learning approach for fine-grained NER, enabling effective entity recognition with minimal annotated training data. Our method can be applied to software domain tasks such as information retrieval, question answering, and article summarization.
 
ine-grained NER \cite{mai2018fine} means classifying entities into more specific classes. And sometimes, the structure of labeled data is heuristic. Therefore, it increases the difficulty of annotating entities. Given the high cost of manual labeling, few-shot learning emerges as a practical solution. By training models with minimal annotated examples, we achieve accurate and efficient fine-grained NER.
In this paper, we present a study that investigates the Few-shot name entity reorganization in the software-related domain. We demonstrate the effectiveness of our method on the StackOverflow dataset \cite{codener}. We proposed a software-related Few-shot NER model that uses the attention network to extract corpus-level information from the code snippets and generate the very first results when identifying 20 types of software-related named entities. Our contributions include: 

\begin{itemize}
\item We propose an entity label interpretation module with a few-shot instance for software-related entities
\item On the StackOverflow dataset we generate the very first result
\end{itemize}

\section{Related Works}

Several studies have been done on the software knowledge base. For example, finding better quality measurement of a question in StackOverflow \cite{ravi}, finding related questions and answers in StackOverflow \cite{shirani}. However, their research lacks NLP techniques for identifying software-related name entities with natural language corpus.

There also several works have been done in NER, \cite{Li19, li2023deception} studied information retrieval and ontology on a specific domain. \cite{qun} propose a domain portable zero-shot learning for NER in task-oriented agents. Recently, deep learning has become popular in NER, especially with self-supervised pre-trained language models (PLMs) such as BERT \cite{bert} and RoBERTa \cite{roberta}. Despite PLMs, NER is a labor-intensive, time-consuming task, which requires domain knowledge experts to annotate a large corpus of domain labels as a train and test set to make the model well. Several studies have been done in Few-shot Fine-Grained NER. \cite{fgner} develop an effective prompt-learning pipeline to predict the type of entities. Another method is using a type-based contextualized instance generator to enlarge the training set and use KL divergence to calculate the loss of newly generated instances \cite{fewshotner}. \cite{liu2023influence} use a discovery algorithm that leverage relevance of the published works. \cite{pmlr-v202-zeng23c}, \cite{zeng23acm} use generative graph learning to find the relevance of entity. However, to our knowledge, no work has been done on software in-domain Few-shot Fine-Grained NER.

There is also a fine-grained NER corpus for the computer programming domain \cite{codener}. The goal of StackOverflow NER is to identify software in-domain name entities. They trained in-domain Bert and combined it with contextual word embeddings and domain-specific embeddings. 

\paragraph{Prompt Learning}
Prompt Learning transfers traditional NLP tasks into prediction problems. The model needs to predict information for unfilled slots. For few-shot entity typing tasks, \cite{fgner}, \cite{fewshotner} add a prompt template containing the entity mentioned after the original sentence.

\paragraph{Meta Learning}
Meta-Learning describes as “learn to learn” While training good model requires abundant labeled data, for the in-domain task, annotations are limited. Meta-learning method can enable model adaptation or generalization to noval tasks encountered during training. \cite{Decomposed} applied meta-learning to Few-Shot NER. They initialized entity detection models using model-agnostic meta-learning (MAML) and proposed MAML-ProtoNet for few-shot entity typing, bridging domain gaps.

\section{Method}
To our best knowledge, there is no few-shot entity typing research that has been done on this StackOverflowNER dataset. Tabassum explored the supervised method on this corpus \cite{codener}. However, there are several limitations. First, for in-domain tasks, it is labor-intensive, time-consuming, we usually need domain experts to do annotation. Second, it takes them more than 1 month to train in-domain BERT on 152 million sentences from StackOverflow.
We explore two models to do few-shot learning on this corpus. The first one is the prompt-based fine-tuning model. Second one is to add a meta-learning method for domain-specific tasks.

\begin{figure}[t!]
    \centering
    \includegraphics[width=0.4\textwidth]{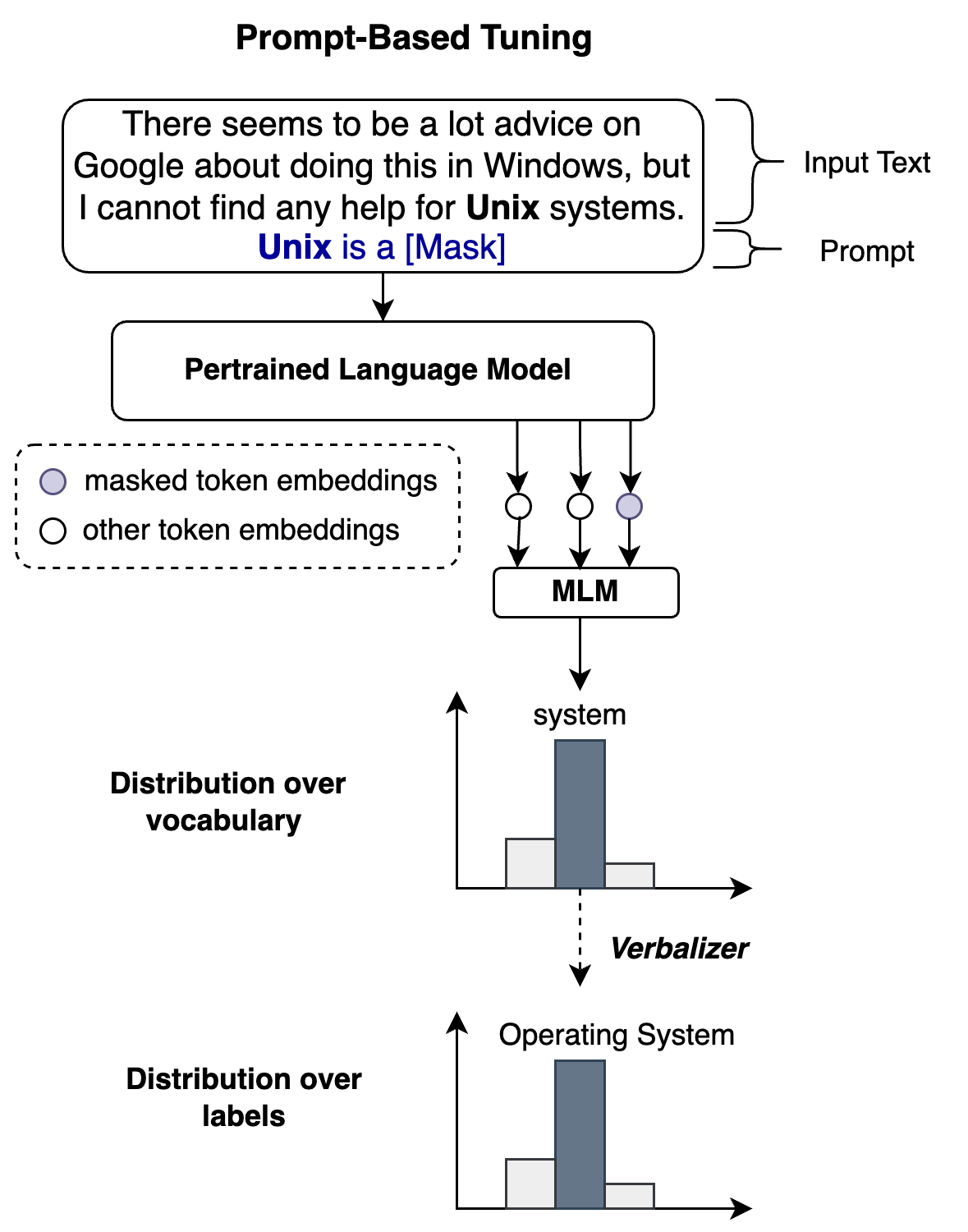}                                                   
    \caption{Prompt based tuning}
    \label{fig:prompt}
\end{figure}

\subsection{Few-shot Entity Typing}
In this section, we will define the problem of Few-shot Entity Typing, which means identifying the type of entities in the sentence with few-shot training data. The input is a sequence of text tokens,$\textbf{x}=\{t_{1},t_{2},...,\textbf{m},...,t_{T}\}$, where $m = \{t_{i},...,t_{j}\}$ is an entity sequence contains $(j-i+1)$ tokens, T is the total number of tokens in the sentence. The output is the label of the entity type $y \in Y$, $ Y $  is the label set $\{y_{1},...,y_{n}\}$, indicating n categories. $K$-shot learning means there are $K$ training examples for each category. We can denote the training dataset as Equation \ref{eqn1}, where $\textbf{m}$ is a $[MASK]$.

\begin{equation} 
\label{eqn1}
\mathcal{D} = \{ (x_{i}, \textbf{m}_{i}, y_{i} )\}_{i=1}^{\left|\mathcal{Y} \right| \times K}
\end{equation}

\subsection{Prompt Based Fine-tuning}
We choose the method proposed by Huang as our baseline model\cite{fewshotner}. The framework of prompt based fine-tuning is shown in Figure \ref{fig:prompt}. Traditional supervised learning trains a model by using given input $x$ to predict the output $y$ as $P(y|x)$. For prompt learning, we need a template that owns unfilled information. We usually use $[MASK]$ token to represent an unfilled slot. For example, a valid template can be as Equation \ref{eqn2}, 

\begin{equation} 
\label{eqn2}
T_{c}(\textbf{x}, \textbf{m}) = \textbf{x}
\end{equation}

After adding a template, the classification problem becomes a prediction problem. Input the sentence with the corresponding template into pre-trained language model encoder $\theta_{0}$ such as RoBERTa. We can get the contextualized representation $h_{m}$ for the $[MASK]$ token as shown in Equation \ref{eqn3},

\begin{equation} 
\label{eqn3}
\textbf{h} = f_{\theta_{0}(x)}
\end{equation}

We need to choose what word can take replace of $[MASK]$ token. By taking the contextualized representation $h_{m}$, the Masked Language Model head is able to get the probability distribution over the entire vocabulary $\mathcal{V}$. We can use the Softmax function to normalize the distribution of words. We can get the probability of a word by giving the contextualized representation $h_m$ of $[MASK]$ as shown in Equation \ref{eqn4}, 

\begin{equation} 
\label{eqn4}
p(w|\textbf{h}) = Softmax(E\sigma(W_{1}\textbf{h} + \textbf{b}_{1}))
\end{equation}

where $E \in \mathbb{R}^{\left| V \right| \times h}$ is the embedding matrix; $\sigma(\cdot)$ is the activation function; $W_{1} \in \mathbb{R}^{h \times h}$ and $b_{1} \in \mathbb{R}^{h}$are from pre-trained Masked Language Model.

After getting the probability of each word w in the vocabulary $\mathcal{V}$. We use a verbalizer to map the prediction over vocabulary to the prediction over labels. We can compute the probability of each label given the probability of each word in the vocabulary as shown in Equation \ref{eqn5},

\begin{equation} 
\label{eqn5}
p(y|\textbf{x}) = \frac{1}{\left| \mathcal{V}_{y} \right|} \sum_{w_{i} \in \mathcal{V}_{y}} \eta_{i}p(w_{i}|\textbf{h})
\end{equation}

We use KL-Divergence as our loss function. The training objective is minimizing the KL-Divergence loss as shown in \ref{eqn6},

\begin{equation} 
\label{eqn6}
D_{KL} (p(Y^{pre}|x) || p(Y^{target}|x))
\end{equation}

where $Y^{pre}$ is the distribution over labels, and $Y^{target}$ is one-hot encoding of target label.

\subsection{Model-Agnostic Meta-Learning}
To improve the performance for the domain-specific task, we add a meta-learning module. The general philosophy of meta-learning is that the model is trained on multitasks to get better parameter initialization. Therefore, the model is able to make quick progress on new domain tasks. Therefore, we apply Model-Agnostic Meta-Learning (MAML) algorithm \cite{finn2017model} in our task. The meta-learning procedure consists of two parts: meta-training on $\xi_{train}$ and meta-testing on $\xi_{test}$. A common expression $K$-shot-$N$-way learning means there are N categories and each category contains $K$ examples. For meta-learning, $K$ should be the same in the meta-training task and meta-testing task. And for better generalization ability, $N$ in the meta-training task should be the same or larger than that in the meta-testing task.

\begin{figure}[t!]
    \centering
    \includegraphics[width=0.4\textwidth]{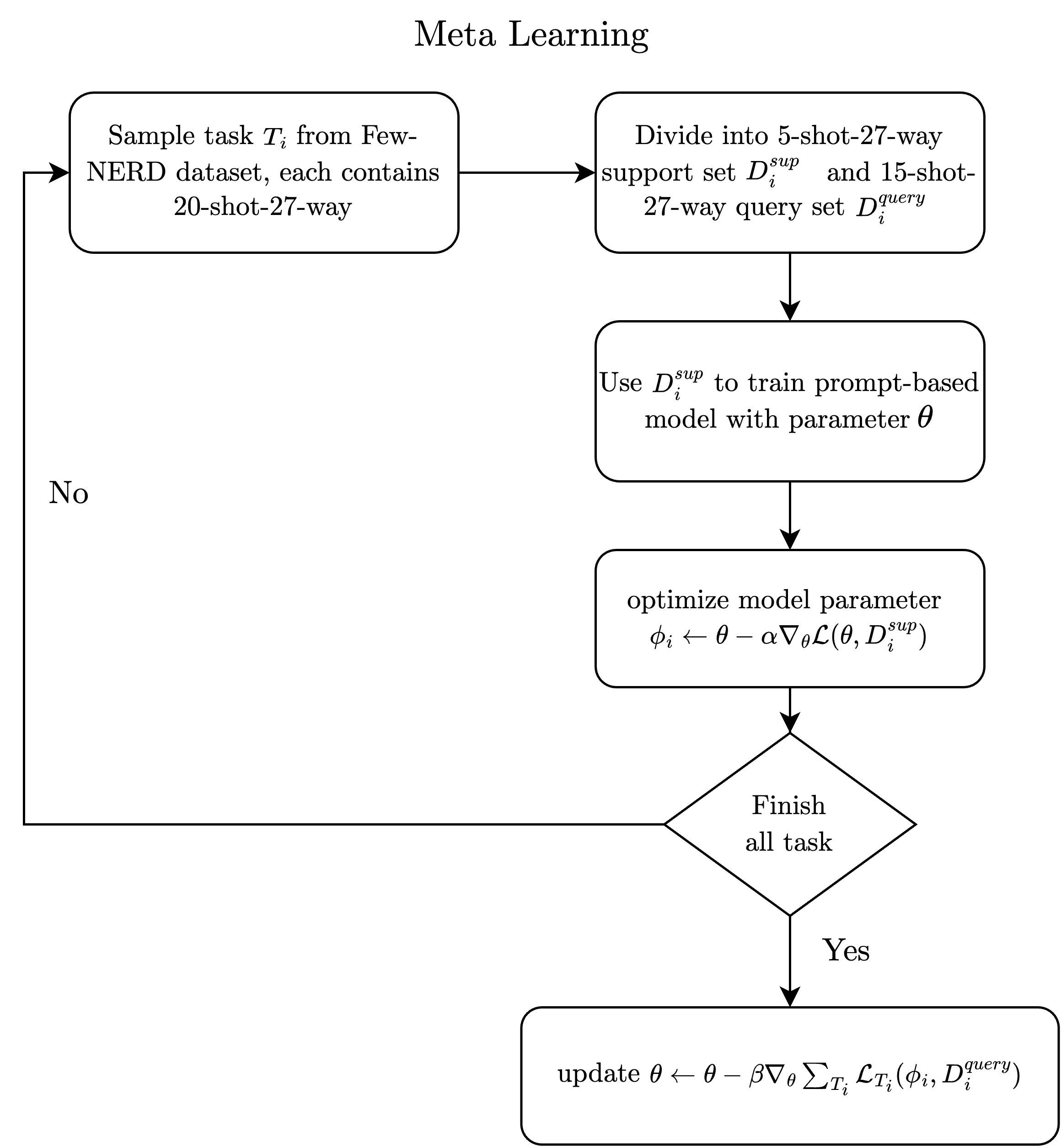}                                                   
    \caption{Meta Training}
    \label{fig:metatraining}
\end{figure}

\subsubsection{Meta-Training}
In the meta-training phase, the model is trained in the general domain dataset. The process of meta-training is shown in Figure \ref{fig:metatraining}. Specifically, we finetune the model on N tasks. For each task, we sample $(D_{i}^{sup} , D_{i}^{query})$ from  $\xi_{train}$ and perform inner update as shown in Equation \ref{eqn7},

\begin{equation} 
\label{eqn7}
 \phi_{i} \leftarrow \theta - \alpha \nabla_{\theta} \mathcal{L}(\theta, D_{i}^{sup})
\end{equation}

where $\alpha$ is the inner update learning rate

For better computing speed, we use a one-step gradient update. We then evaluate $\phi_{i}$ on the $D_{i}^{query}$ and execute meta-update by aggregating losses in all tasks as shown in Equation \ref{eqn8},

\begin{equation} 
\label{eqn8}
\sum_{T_{i}} \mathcal{L}_{T_{i}}(\phi_{i}, D_{i}^{query})
\end{equation}

and update the model parameter $\nabla_{\theta}$ as the follow Equation \ref{eqn9},

\begin{equation} 
\label{eqn9}
  \theta \leftarrow \theta - \beta \nabla_{\theta} \sum_{T_{i}} \mathcal{L}_{T_{i}}(\phi_{i}, D_{i}^{query})
\end{equation}

where $\beta$ is the learning rate of meta-learning.

\begin{figure}[t!]
    \centering
    \includegraphics[width=0.3\textwidth]{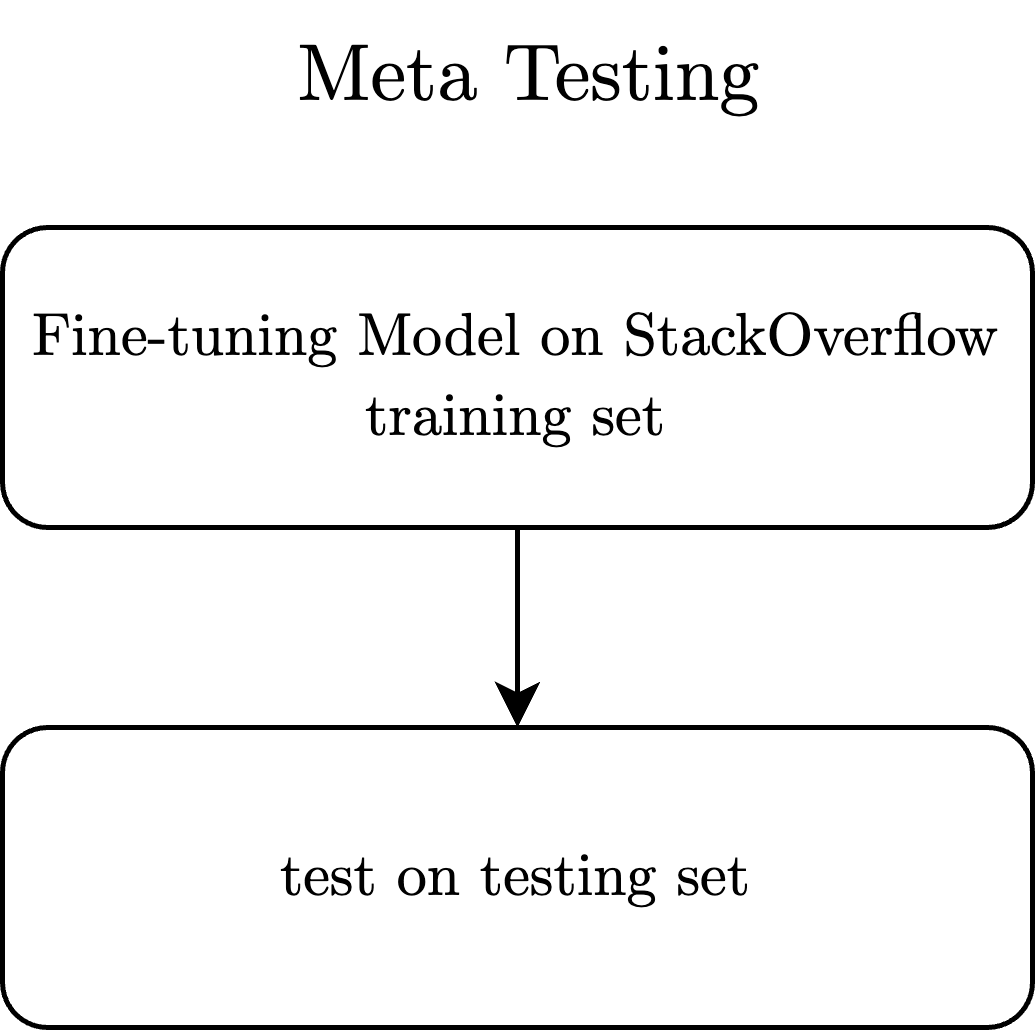}                                                   
    \caption{Meta Testing}
    \label{fig:metatesting}
\end{figure}

\subsubsection{Meta-Testing}
In meta-testing, we use an updated parameter to finetune the model on the StackOverflow training dataset and make a prediction on the testing dataset. The process of meta-testing is shown in Figure \ref{fig:metatesting}.

\section{Experiment}

\subsection{Dataset}

\begin{table}[t!]
\centering
\scalebox{0.9}{
\begin{tabular}{ll}
\hline
\textbf{Input} & \textbf{Category} \\
\hline
{[}JSP{]} code is .                            & Library         \\ \hline
Thanks, {[}AJ{]}                       & User Name     \\ \hline
This is my part of my {[}xml{]} file   & File Type     \\ \hline
Code to implement the {[}WindRose{]} : & Class Name    \\ \hline
Ditto for JmenuBar, {[}JMenu{]}.       & Library Class \\ \hline
I am working  on {[}SLES{]} 11 .             & OS            \\ \hline
Your {[}main{]} method then becomes    & Function Name \\ \hline
Use {[}tab{]} during this process.     & Keyboard IP   \\ \hline

\hline
\end{tabular}}
\caption{Dataset Example}
\label{table:dataexp}

\end{table}

In this task, we will use StackOverflow NER corpus \cite{codener}, which contains more than 1,237 question-answer threads from StackOverflow 10-year archive with 27 types of entities. 
We are given code entities and natural language entities. Code entities include \emph{Class, Variable, In Line Code, Function, Library, Value, Data Type}, and \emph{HTML XML Tag}. Natural language entities include \emph{Application, UI Element, Language, Data Structure, Algorithm, File Type, File Name, Version, Device, OS, Website}, and \emph{User Name}. We also plan to use additional Github data which randomly sampled repositories from GitHub. An example of our dataset and entity labels is shown in table \ref{table:dataexp}.

\subsection{Experiment Settings}
\textbf{Few-Shot Sampling.} We conduct 5-shot learning on StackOverflow NER corpus by sampling 5 instances for the training set in each run of the experiment. We also did an experiment in which we manually select 5 instances over the entire training set to make the model prediction more accurate. 

\textbf{Meta-Training Sampling.} Few-NERD dataset contains 66 fine-grained entity types in the general domain. We perform meta-training phase in 40 tasks. For each task, we randomly sample 20-shot-27-way dataset from Few-NERD. And then we split it into 5-shot-27-way support set and 15-shot-27-way query set. 

\textbf{Hyperparameter Settings.} We use the pre-trained RoBERTa-base model as the fundamental model. The max sequence length is 128; the inner batch size for fine-tuning is 8; the outer batch size for meta-updating is 32; the finetuning epoch number in meta-training is 1; the finetuning epoch number in meta-testing is 10; the maximum meta-learning step is 15; the learning rate for meta-learning is 5e-3; the inner finetuning learning rate is 1e-2. We trained the model on Google Colab GPU. 

\textbf{Evaluation Metrics.} We apply the loose micro-F1 score (micro-F1) and loose macro-F1 score(macro-F1)

\subsection{Results}

\begin{table}
\centering
\begin{tabular}{lll}
\hline
\textbf{Model} & \textbf{Micro-F1} & \textbf{Macro-F1}\\
\hline
RoBERTa & 0.3091 & 0.2831 \\
BertOverflow & N/A & N/A \\
RoBERTa+MAML & 0.3578 & 0.3197 \\

\hline
\end{tabular}
\caption{
Performance comparision between baseline model RoBERTa and RoBERTa+MAML
}
\label{citation-guide}
\end{table}

We applied RoBERTa and RoBERTa+MAML to the StackOverflow NER Corpus. For the training set, we randomly select five samples for each category from the StackOverFlow NER Corpus as the input data. As the result is shown in Table \ref{citation-guide}, the Micro-F1 score of the baseline model RoBERTa is 0.3091 and the Macro-F1 score is 0.2837. 

We also apply the same method for our RoBERTa+MAML model. The Micro-F1 score of the baseline model RoBERTa+MAML is 0.3578 and the Macro-F1 score is 0.3197. We observe a large increase in the Micro-F1 score by using the model with meta-learning. 

\begin{figure}[t!]
    \begin{subfigure}{\linewidth}
        \centering
        \includegraphics[width=0.9\linewidth]{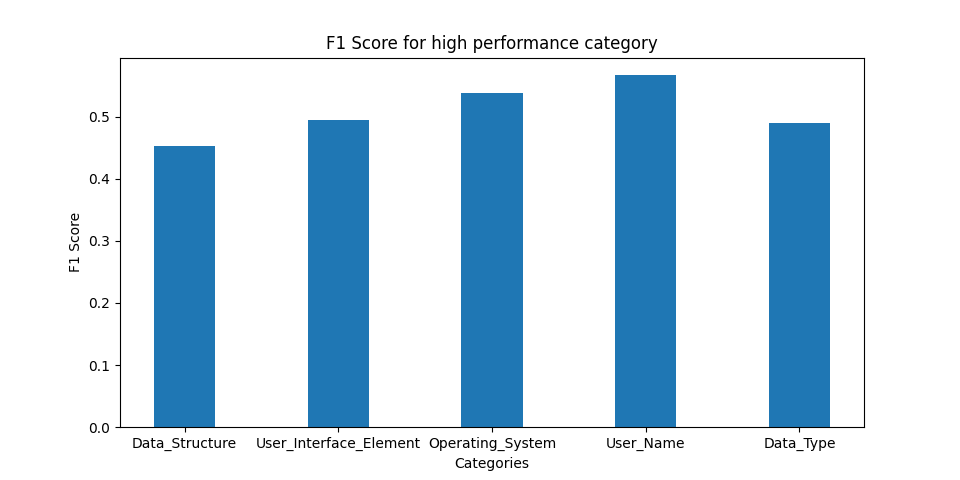}                                                  
        \caption{Categories with Higher F1 Score}
        \label{fig:highperform}
    \end{subfigure}
    \begin{subfigure}{\linewidth}
        \centering
        \includegraphics[width=0.9\linewidth]{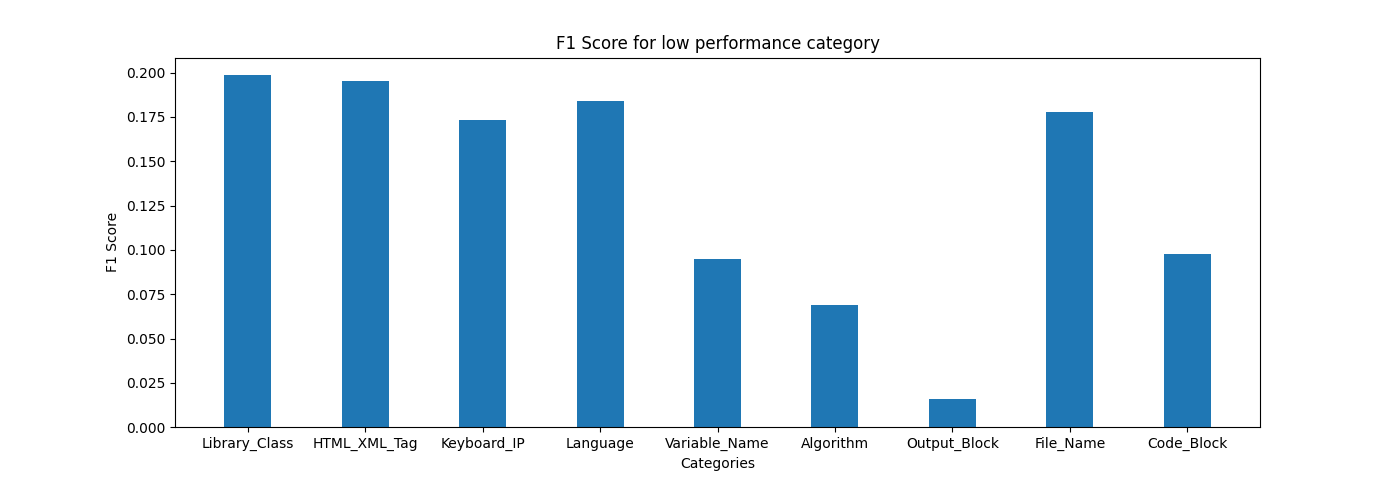}                                                  
        \caption{Categories with Lower F1 Score}
        \label{fig:lowperform}        
    \end{subfigure}
    \caption{Different categories F1 Score}
    \label{fig:f1compare}
\end{figure}

\begin{figure}[t!]
    \begin{subfigure}{\linewidth}
        \centering
        \includegraphics[width=0.9\linewidth]{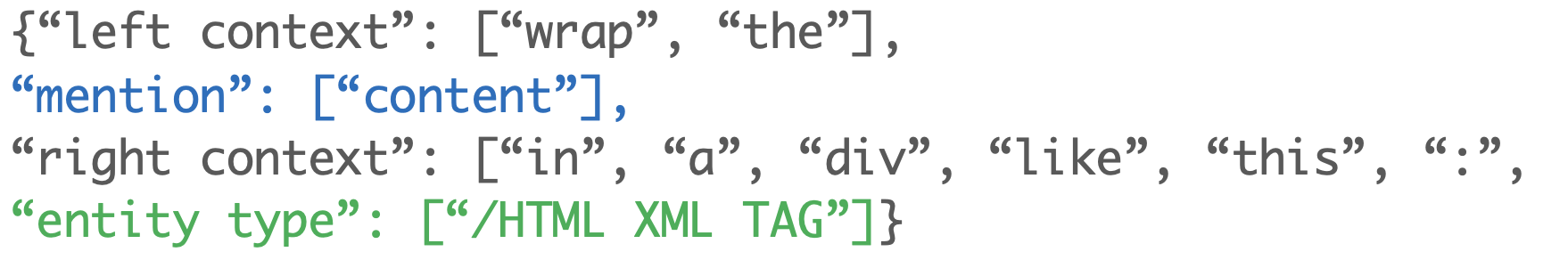}      
        \caption{Example for an appropriate training data}
         \label{fig:good}
    \end{subfigure}
    \begin{subfigure}{\linewidth}
        \centering
        \includegraphics[width=0.9\linewidth]{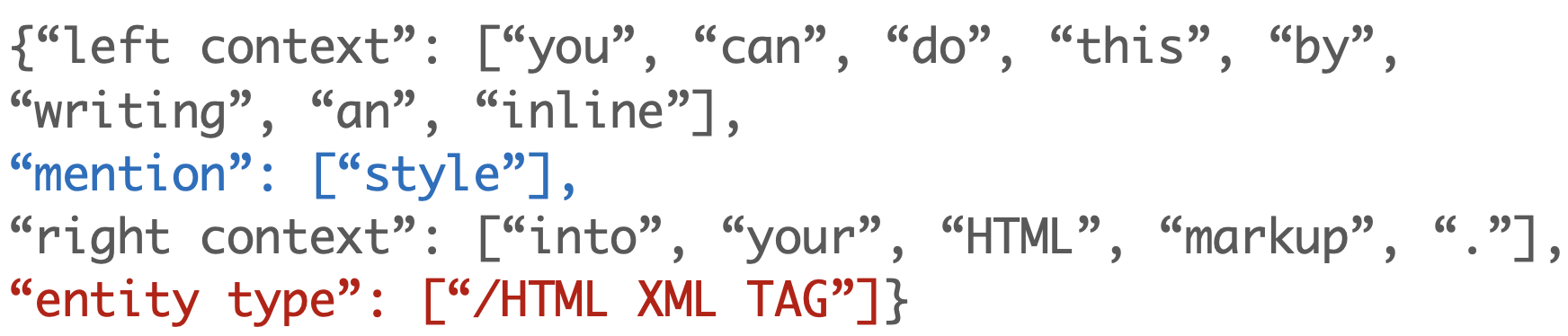}       
        \caption{Example for an inappropriate training data}
         \label{fig:bad}
    \end{subfigure}
    \caption{Example between different training data}
    \label{fig:compare}
\end{figure}

As we can see in Figure (\ref{fig:highperform}, \ref{fig:lowperform}), \textit{Data structure, User Interface Element, Operating System, User Name}, and \textit{Data Types} are categories that RoBERTa+MAML can recognize better.

\subsection{Case Study for 5-Shot Training Set}

\begin{table}
\centering
\begin{tabular}{lll}
\hline
\textbf{Model} & \textbf{Micro-F1} & \textbf{Macro-F1}\\
\hline
Original Input & 0.3254 & 0.2986 \\
Processed Input & 0.3578 & 0.3197 \\
\hline
\end{tabular}

\caption{
Performance comparison between 5-shot training data processing and after-training data processing
}
\label{table:beforeafter}
\end{table}

We notice that the F1 score of many categories (e.g. \textit{OS, Library Class, Function Name, Keyboard IP, Language, Variable Name}, and \textit{Algorithm} ) are lower than usual. Those categories are difficult to recognize in the 5-shot domain-specific NER learning. After exploring our randomized training set, we found that the training set contains many duplicate entities and some of the entities are ambiguous.  

In order to avoid the negative impact of the 5-shot training dataset, we manually choose the meaningful and representative training data. We make each training data unique and representative.

We also apply the same method for our RoBERTa+MAML model. As shown in Table \ref{table:beforeafter}, the Micro-F1 score is improved by around 3\% and the Macro-F1 score is improved by around 2\%.

\begin{figure}[t!]
    \begin{subfigure}{\linewidth}
        \centering
        \includegraphics[width=0.9\linewidth]{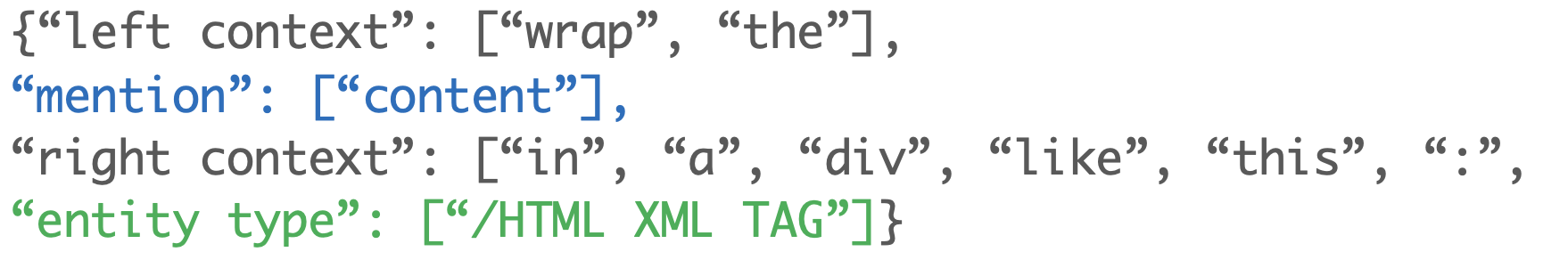}      
        \caption{Example for an appropriate training data}
         \label{fig:good}
    \end{subfigure}
    \begin{subfigure}{\linewidth}
        \centering
        \includegraphics[width=0.9\linewidth]{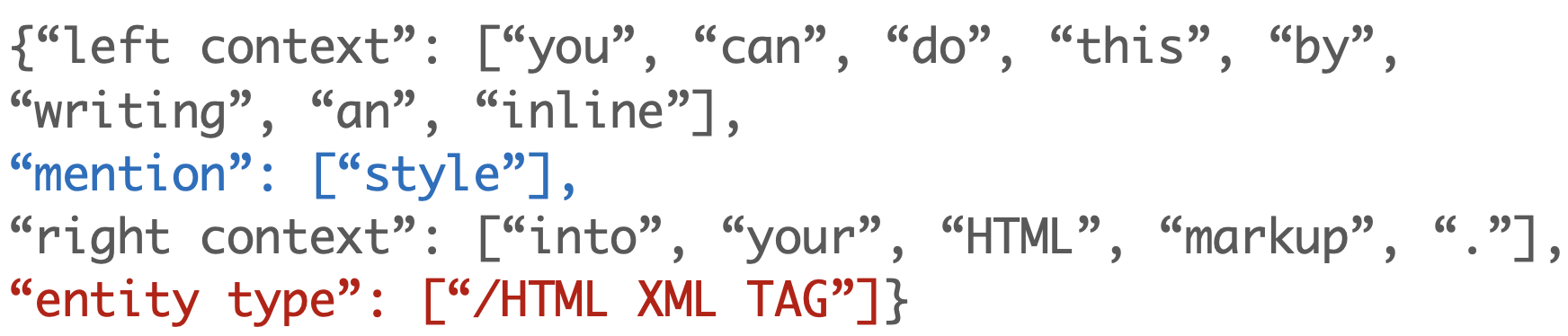}       
        \caption{Example for an inappropriate training data}
         \label{fig:bad}
    \end{subfigure}
    \caption{Example between different training data}
    \label{fig:compare}
\end{figure}

This is an example shows that ``content" is ambiguous (Figure \ref{fig:good}, \ref{fig:bad}), which does not contain the generality in the 5-shot training data. Thus, we manually select the substitute training data in order to maximize the model's performance. According to the result, RoBERTa+MAML with manually selected training data outperforms the randomized training data. Thus, we adopt the manually selected training data for future work. 

\subsection{Case Study for Knowledge Based Pattern Extraction }

\begin{table}
\centering
\scalebox{0.9}{
\begin{tabular}{|l|l|}
\hline
Data & Category  \\
\hline
 I have problem opening {[}CVS{]} from PHP code & \multirow{4}{*}{File Type}  \\
I attached my {[}YUV\_420\_888{]} to the problem & \\
How to open {[}XLSX{]} file locally & \\
How can I use the dll and the header file & \\ \hline
 It is a correct format ... I am using {[}windows{]} 10 & \multirow{2}{*}{OS} \\
I am making an {[}android{]} application & \\ \hline
\end{tabular}
}
\caption{
Samples for common patterns. We can use knowledge-based pattern extraction to handle certain categories, such as File types and OS. }
\label{table:maual}
\end{table}

\begin{table}
\centering
\scalebox{0.9}{
\begin{tabular}{llll}
\hline
\textbf{Model} & \textbf{F1} & \textbf{Precision} & \textbf{Recall}\\
\hline
RoBERTa+MAML & 0.345 & 0.402 & 0.302 \\
Knowledge Based Pattern Extraction & 0.490 & 0.716 & 0.372 \\
\hline
\end{tabular}}
\caption{
Performance comparison between ROBERTa+MAML and knowledge-based pattern extraction}
\label{table:extraction}
\end{table}

We notice that certain categories are difficult to recognize but contain patterns. For example, most file types exist within the common file type collections. As shown in Table \ref{table:maual}, we can perform a knowledge-based pattern extraction to extract certain categories. For example, we use regular expressions to extract common file name extensions, such as csv, jpg, and doc. As shown in Table \ref{table:extraction}, the F1 score for the file type category is largely improved from 0.345 to 0.49. Precision and Recall scores arrive at 0.716 and 0.372. We can also apply knowledge-based pattern extraction to many other categories, which will greatly improve the overall prediction accuracy and F1 score.

\section{Conclusion \& Future Work}

In our study, we focused on domain-specific NER within the computer programming domain. Leveraging RoBERTa + MAML on the 5-shot StackOverflow NER corpus, we observed substantial improvements over the baseline RoBERTa model. Meta-learning comes as a powerful tool for few-shot domain-specific NER tasks. Additionally, domain-specific phrase processing and knowledge-based pattern extraction further enhanced accuracy. We anticipate that meta-learning, domain-specific phrase processing, and knowledge-based patterns will benefit future software-related information extraction and question-answering tasks. To refine our approach, we aim to expand our dataset variety and explore additional sample support sets and query sets, thereby amplifying the impact of meta-learning.

\bibliographystyle{IEEEtran}
\bibliography{IEEEtran}
\end{document}